\documentclass{article}
\usepackage{ijcai23}

\usepackage{times}
\usepackage{soul}
\usepackage{url}
\usepackage[hidelinks]{hyperref}
\usepackage[utf8]{inputenc}
\usepackage[small]{caption}
\usepackage{graphicx}
\usepackage{amsmath}
\usepackage{amsthm}
\usepackage{booktabs}
\usepackage{algorithm}
\usepackage[switch]{lineno}

\title{SecureBoost Hyperparameter Tuning via Multi-Objective Federated Learning}

\author{
Ziyao Ren$^1$\and
Yan Kang$^{2 \dagger}$\and
Lixin Fan$^2$\and\\
Linghua Yang$^1$\and
Yongxin Tong$^1$\And
Qiang Yang$^{2,3}$
\affiliations
$^1$State Key Laboratory of Software Development Environment, Beijing Advanced Innovation Center for Future Blockchain and Privacy Computing, School of Computer Science, Beihang University\\
$^2$Webank\\
$^3$Hong Kong University of Science and Technology
\emails
\{ziyaoren, larryhawkingyoung, yxtong\}@buaa.edu.cn,
\{yangkang, lixinfan\}@webank.com,\\
qyang@cse.ust.hk
}

\usepackage{afterpage}
\usepackage{makecell}
\usepackage{tabulary}
\usepackage{xcolor}
\usepackage{colortbl}
\usepackage{prettyref}
\usepackage{framed}
\usepackage{booktabs}       

\newcommand{\savehyperref}[2]{\texorpdfstring{\hyperref[#1]{#2}}{#2}}
\newrefformat{eq}{\savehyperref{#1}{Eq. \textup{(\ref*{#1})}}}
\newrefformat{eqn}{\savehyperref{#1}{Eq.~\textup{(\ref*{#1})}}}
\newrefformat{lem}{\savehyperref{#1}{Lemma~\ref*{#1}}}
\newrefformat{assump}{\savehyperref{#1}{Assumption~\ref*{#1}}}
\newrefformat{defi}{\savehyperref{#1}{Definition~\ref*{#1}}}\newrefformat{tab}{\savehyperref{#1}{Table~\ref*{#1}}}
\newrefformat{lemma}{\savehyperref{#1}{Lemma~\ref*{#1}}}
\newrefformat{line}{\savehyperref{#1}{Line~\ref*{#1}}}
\newrefformat{thm}{\savehyperref{#1}{Theorem~\ref*{#1}}}
\newrefformat{corr}{\savehyperref{#1}{Corollary~\ref*{#1}}}
\newrefformat{cor}{\savehyperref{#1}{Corollary~\ref*{#1}}}
\newrefformat{sec}{\savehyperref{#1}{Section~\ref*{#1}}}
\newrefformat{app}{\savehyperref{#1}{Appendix~\ref*{#1}}}
\newrefformat{ex}{\savehyperref{#1}{Example~\ref*{#1}}}
\newrefformat{fig}{\savehyperref{#1}{Figure~\ref*{#1}}}
\newrefformat{alg}{\savehyperref{#1}{Algorithm~\ref*{#1}}}
\newrefformat{rem}{\savehyperref{#1}{Remark~\ref*{#1}}}
\newrefformat{conj}{\savehyperref{#1}{Conjecture~\ref*{#1}}}
\newrefformat{prop}{\savehyperref{#1}{Proposition~\ref*{#1}}}
\newrefformat{proto}{\savehyperref{#1}{Protocol~\ref*{#1}}}
\newrefformat{prob}{\savehyperref{#1}{Problem~\ref*{#1}}}
\newrefformat{claim}{\savehyperref{#1}{Claim~\ref*{#1}}}
\newrefformat{que}{\savehyperref{#1}{Question~\ref*{#1}}}
\newrefformat{op}{\savehyperref{#1}{Open Problem~\ref*{#1}}}
\newrefformat{fn}{\savehyperref{#1}{Footnote~\ref*{#1}}}
\newrefformat{property}{\savehyperref{#1}{Property~\ref*{#1}}}

\usepackage{xspace}
\usepackage{amsmath}
\usepackage{amsthm}
\usepackage{bbm}
\usepackage[noend]{algpseudocode}
\usepackage{mathrsfs}
\usepackage{amssymb}
\usepackage{bm}
\usepackage{makecell}
\usepackage{tabulary}
\usepackage{subcaption}

\newcommand{\gray}[1]{{\color{gray}#1}}

\newtheorem{definition}{Definition}

\newtheorem{remark}{Remark}

\theoremstyle{definition}

\newcommand{\fakeparagraph}[1]{\vspace{1mm}\noindent\textbf{#1.}}

\begin{document}

\maketitle
\def\thefootnote{\arabic{footnote}}
\def\thefootnote{$\dagger$}\footnotetext{Corresponding author}\def\thefootnote{\arabic{footnote}}
\begin{abstract}
  SecureBoost is a tree-boosting algorithm leveraging homomorphic encryption to protect data privacy in vertical federated learning setting. It is widely used in fields such as finance and healthcare due to its interpretability, effectiveness, and privacy-preserving capability. However, SecureBoost suffers from high computational complexity and risk of label leakage. To harness the full potential of SecureBoost, hyperparameters of SecureBoost should be carefully chosen to strike an optimal balance between utility, efficiency, and privacy. Existing methods either set hyperparameters empirically or heuristically, which are far from optimal. To fill this gap, we propose a Constrained Multi-Objective SecureBoost (CMOSB) algorithm to find Pareto optimal solutions that each solution is a set of hyperparameters achieving optimal tradeoff between utility loss, training cost, and privacy leakage. We design measurements of the three objectives. In particular, the privacy leakage is measured using our proposed instance clustering attack. Experimental results demonstrate that the CMOSB yields not only hyperparameters superior to the baseline but also optimal sets of hyperparameters that can support the flexible requirements of FL participants. 
\end{abstract}

\section{Introduction}

Federated learning~\cite{DBLP:conf/aistats/McMahanMRHA17,DBLP:journals/debu/TongZZLSL0L23} is a novel distributed machine learning paradigm that enables multiple participants to train machine learning models using participants' private data without compromising data privacy. 
SecureBoost~\cite{DBLP:journals/expert/ChengFJLCPY21} is a popular vertical federated gradient boosting tree algorithm for its interpretability and effectiveness. It has been widely applied in many fields, such as advertisement, finance, and healthcare. 

SecureBoost applies homomorphic encryption (HE)~\cite{DBLP:conf/eurocrypt/Paillier99} to protect data privacy. 
However, HE is computationally expensive, which hinders SecureBoost from being applied to large-scale machine learning problems. 
In addition, we demonstrate that SecureBoost suffers from the instance clustering attack (ICA) that we propose to infer labels of the active party. In other words, SecureBoost has the risk of leaking labels (see Sec.~\ref{sec:priv}). We design two countermeasures against ICA but at the expense of losing utility.

To harness the full potential of SecureBoost, its hyperparameters (including hyperparameters of the applied protection mechanism) should be set such that the utility loss, training cost, and privacy leakage can be minimized simultaneously, and the optimal tradeoffs between utility, efficiency, and privacy can be achieved. However, existing FL platforms~\cite{DBLP:journals/jmlr/LiuFCXY21,DBLP:conf/sigmod/FuSYJXT021} often set the hyperparameters of SecureBoost to empirical values or apply heuristic methods to search for the hyperparameters to tradeoff between utility, efficiency, and privacy. These existing methods may lead to hyperparameters far from satisfactory in practice. These gaps inspire us to investigate the following critical open problem: \textit{is it possible to design a systematic approach that can find SecureBoost hyperparameters to achieve optimal tradeoffs between utility, efficiency, and privacy?}

Our work provides a positive answer to this question. More specifically, we apply Constrained Multi-Objective Federated Learning (CMOFL)~\cite{kang2023optimizing} to find Pareto optimal solutions of hyperparameters that can minimize utility loss, training cost, and privacy leakage simultaneously. Each solution is an optimal tradeoff between the three conflicting objectives. As a result, Pareto optimal solutions not only provide optimal hyperparameters for SecureBoost but also can support the flexible requirements of FL participants. For example, FL participants can select the most appropriate hyperparameters from the Pareto optimal solutions that satisfy their current preference over utility, efficiency, and privacy.

Our main contributions are summarized as follows: 
\begin{itemize}
    \item We apply Constrained Multi-Objective Federated Learning (CMOFL) to SecureBoost and propose a Constrained Multi-Objective SecureBoost (CMOSB) algorithm (Sec.~\ref{sec:mosb}) to find the Pareto optimal solutions of hyperparameters that minimize utility loss, training cost, and privacy leakage simultaneously. This is the first attempt to apply CMOFL to VFL in general and SecureBoost in particular. 
    \item We design measurements of utility loss, training cost, and privacy leakage for SecureBoost (Sec. \ref{sec:objectives}). In particular, privacy leakage is measured by our proposed label inference attack. We also propose two countermeasures against this attack (Sec.~\ref{sec:priv}).
    \item We conduct experiments on four datasets, demonstrating that our CMOSB can find far better hyperparameters than default ones set by FATE~\cite{DBLP:journals/jmlr/LiuFCXY21} and VF\textsuperscript{2}Boost~\cite{DBLP:conf/sigmod/FuSYJXT021} in terms of the tradeoff between privacy leakage, utility loss, and training cost. Moreover, CMOSB can find Pareto optimal solutions of hyperparameters that achieve an optimal tradeoff between utility loss, training cost, and privacy leakage (Sec.~\ref{sec:exp}).
\end{itemize}

\section{Related Work}\label{sec:rel}

\subsection{Label Leakage in VFL}

Label leakage in vertical federated learning refers to the situation where the passive party is able to obtain the label data of the active party during the training process. 
\cite{DBLP:conf/uss/Fu0JCWG0L022} systematically classified the label leakage in vertical federated learning into three categories: active attack, passive attack, and direct attack, and provided an attack method for each category.
\cite{liu2021batch} proposed a label attack method on the black-boxed VFL and presented a privacy protection method. 
\cite{DBLP:journals/corr/abs-2301-07284} demonstrated that split learning remains vulnerable to label inference attacks.
\cite{LabelRelation22} finds that the training of GNNs in VFL may also lead to the leakage of sample relationships. 
\cite{DBLP:conf/bigcom/TanZLLW22} proposed a gradient inversion attack that utilizes the gradients of local models to reconstruct label data.

\subsection{Multi-objective Federated Learning}

Multi-objective federated learning is a collaborative optimization approach in which participants of FL aim to optimize multiple conflicting objectives simultaneously and find Pareto optimal solutions. 
\cite{DBLP:conf/nips/CuiPLZW21} considered the utility of participating parties as the optimization objective, model disparity as the constraint, and optimized the objectives of all participants to calculate the Pareto front. 
This approach can ensure the fairness of federated learning and obtain the optimal performance model. \cite{DBLP:journals/tnn/ZhuJ20} formulated the accuracy and communication cost of federated learning as the optimization objectives and adjusted the model sparsity to minimize both the communication cost and test errors.
The algorithm proposed in \cite{kang2023optimizing} is a multi-objective federated learning algorithm with constraints, which add constraints to the optimization objectives and find the Pareto front faster within the feasible range.
\section{Background}\label{sec:bac}

\subsection{Vertical Federated Learning}

\begin{figure}[htbp]
    \vspace{-6pt}
    \centering
    \includegraphics[width=0.99\linewidth]{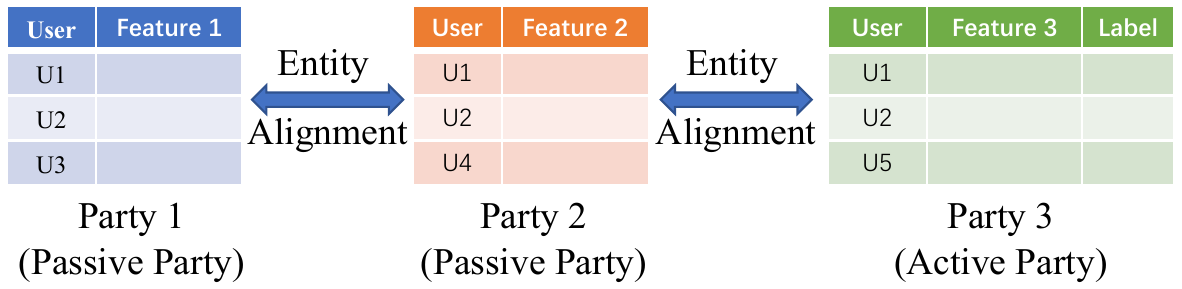}
    \vspace{-6pt}
    \caption{An illustration of data partition in VFL. }
    \label{fig:verticalfed}
    \vspace{-6pt}
\end{figure}

Vertical Federated Learning~\cite{DBLP:journals/tist/YangLCT19} is one of the scenarios in Federated Learning, which refers to the situation where feature data is vertically partitioned among multiple parties, with one party possessing the data labels (Fig.~\ref{fig:verticalfed}): The active party holds features and labels while passive parties hold only features. 

All participating parties in VFL first align their instances by private set intersection (PSI)~\cite{DBLP:conf/eurocrypt/FreedmanNP04}, and then perform federated model training and inferencing. 
During the training process, each party is not allowed to reveal its training data to others~\cite{DBLP:journals/pvldb/TongPZSXZZCXXL22}. 
After the federated training is completed, the federated model $M$ is obtained, which is jointly held by all participating parties, i.e., model $M$ is split into $M_1, M_2, \dots, M_K$. 
In the inferencing phase, all participating parties collaboratively make the prediction but only the active party can access the final predicted result. 

\subsection{SecureBoost}
SecureBoost~\cite{DBLP:journals/expert/ChengFJLCPY21} is a widely used gradient boosting tree algorithm designed for the vertical federated learning scenario. The core idea of SecureBoost is to use $n$ federated decision trees to fit instance labels. 

For each iteration, SecureBoost constructs a new tree from the root node based on a node split finding algorithm, which is summarized in Algo.~\ref{alg:split}: the active party sends the homomorphically encrypted gradients $\left\langle g\right\rangle$, $\left\langle h\right\rangle$, and instance space $I$ of the current node to passive parties, 
who, in turn, calculate the gradient statistics and send them back to the active party; the active party finds the optimal split of the current node with the maximal splitting score and informs the corresponding parties to split the instance space into child nodes. 
The split finding process continues until the maximum depth is reached. We refer readers to \cite{DBLP:journals/expert/ChengFJLCPY21} for a more detailed algorithm description.

\begin{algorithm}[!ht]\vspace{-3pt}
    \caption{SecureBoost Split Finding (SF)}
	\begin{algorithmic}[1]
    \vspace{2pt}
    \Statex \textbf{Input:} Instance space of current node $I$; 
    \Statex \textbf{Input:} Gradient $\left\langle g\right\rangle$, hessian $\left\langle h\right\rangle$.
    \Statex \textbf{Output:} Partition current instance space according to the selected attribute's value. 
    \State \gray{$\triangleright$ \textit{Passive parties perform:}}
  \State {Calculate gradient statistics based on $\left\langle g\right\rangle$, $\left\langle h\right\rangle$, and $I$; }
  
    \State \gray{$\triangleright$ \textit{Active party performs:}}
    \For{each split point}
        \State {Calculate info gain based on gradient statistics;}
        \State {Update the splitting score based on the info gain;}
    \EndFor
    \State {Return optimal split to the corresponding party;}
    \State \gray{$\triangleright$ \textit{Passive party performs:}}
    \State {Partition instance space to form $I_L$ according to the optimal split, and update model;}
    \State \gray{$\triangleright$ \textit{Active party performs:}}
    \State {Split current node according to $I_L$ and update model;}
	\end{algorithmic}\label{alg:split}
\end{algorithm}

The instance distributions of leaf nodes in SecureBoost are not protected by HE, which may lead to privacy leakage. Some methods~\cite{DBLP:journals/pvldb/WuCXCO20,DBLP:conf/cikm/FangZT0YWWZZ21}
use a combination of HE and MPC to protect privacy to address this issue. However, these methods may result in significant training cost, making it challenging to be applied in trustworthy federated learning. 

\subsection{Constrained Multi-Objective Federated Learning}

The Constrained Multi-Objective Federated Learning (CMOFL) problem aims to find Pareto optimal solutions that simultaneously minimize multiple FL objectives under constraints.
We adapt the definition of CMOFL proposed in the work~\cite{kang2023optimizing} to the VFL setting.

\begin{definition}[Constrained Multi-Objective Vertical Federated Learning] The problem of Constrained Multi-Objective Vertical Federated Learning is formulated as follows:\label{def:constra-multi-objective-FL}
\begin{equation}
    \begin{split}
        &\min\limits_{x \in \mathcal{X}} ( f_1(x), f_2(x), \ldots , f_m(x) ) \\
        &\text{where } f_i(x) = \sum_{k=1}^KF_{i,k}(x)\text{ for $1\leq i\leq m$} \\
        & \text{subject to } \,\, f_j(x) \leq \phi_j, \forall j\in \{1,\cdots,m\}
    \end{split}
\end{equation}
where $x \in \mathbb{R}^d$ is a solution in the decision space $\mathcal{X}$, $\{f_i\}_{i=1}^m$ are the $m$ objectives to minimize, $F_{i,k}$ is the local objective of client $k$ corresponding to the $i$th objective $f_i$, and $\phi_i$ is the upper bound of $f_i$.
\end{definition}

\begin{remark}
Each objective in Definition~\ref{def:constra-multi-objective-FL} is the sum of the local objectives of all parties instead of the average defined in \cite{kang2023optimizing} because participants in VFL are integral parts of the whole VFL system.
\end{remark}

The FL participants aim to find Pareto optimal solutions and front for the Constrained Multi-Objective VFL problem. We provide the definitions of Pareto dominance, Pareto optimal solution, Pareto set, and Pareto front as follows.

\begin{definition}[Pareto Dominance]\label{def:pareto_dom}
Let $x_a, x_b \in \mathcal{X}$,  $x_a$ is said to dominate $x_b$, denoted as $x_a \prec x_b$, if and only if $f_i(x_a) \leq f_i(x_b), \forall i \in \{1,\ldots,m\}$ and $f_j(x_a) <  f_j(x_b), \exists j \in \{1,\ldots,m\}$.
\end{definition}

\begin{definition}[Pareto Optimal Solution]\label{def:pareto_sol}
A solution $x^{*} \in \mathcal{X}$ is called a Pareto optimal solution if there does not exist a solution $\hat{x} \in \mathcal{X}$ such that $\hat{x} \prec x^*$.
\end{definition}

A Pareto optimal solution refers to a solution that achieves an optimal tradeoff among different conflicting objectives. The collection of all Pareto optimal solutions forms the Pareto set, while the corresponding objective values for the Pareto optimal solutions form the Pareto front. The definitions of the Pareto set and Pareto front are formally given as follows.

\begin{definition}[Pareto Set and Front]\label{def:pareto_set_front}
The set of all Pareto optimal solutions is called the Pareto set, and its image in the objective space is called the Pareto front.
\end{definition}

We use hypervolume~(HV) indicator~\cite{DBLP:conf/ppsn/ZitzlerK04} as a metric to measure the Pareto front. 
The definition of hypervolume indicator is provided below.

\begin{definition}[Hypervolume Indicator]\label{def:hypervolume}
Let $z = \{z_1,\cdots, z_m\}$ be a reference point that is an upper bound of the objectives $V = \{v_1,\ldots, v_m\}$, such that $v_{i} \leq z_i$, $\forall i \in [m]$. The hypervolume indicator $\text{HV}(V)$ measures the region between $V$ and $z$ and is formulated as:
\begin{equation}
    \text{HV}(V) = \Lambda \left( \left \{ q \in \mathbb{R}^m \big| q \in \prod_{i=1}^{m}[v_i, z_i]  \right \}\right) 
\end{equation}
where $\Lambda(\cdot)$ refers to the Lebesgue measure.
\end{definition}

\section{Label Leakage in SecureBoost}\label{sec:priv}

In this section, we design a label inference attack and explain how this attack infers labels of the active party. We also propose two defense methods against this attack.

\subsection{Threat Model}
\label{sec:threat}
Below we discuss the threat model, including the attacker's objective, capability, and knowledge.

\noindent\textbf{Attacker's objective}. We consider the passive party as the attacker who aims to infer labels owned by the active party.

\noindent\textbf{Attacker's capability}. We assume the attacker is \textit{semi-honest} in the sense that the attacker faithfully follows the SecureBoost protocol but may mount label inference attacks to infer labels of the active party.

\noindent\textbf{Attacker's knowledge}. 
The passive party can access instance distributions of leaf nodes because SecureBoost does not protect this information. 
We also assume that the passive party holds several labeled instances for each class, a similar assumption made in \cite{DBLP:conf/uss/Fu0JCWG0L022}.

\subsection{Label Inference Attack}
\label{sec:attack}

\begin{figure}[htbp]
    \centering
    \vspace{-6pt}
    \includegraphics[width=0.99\linewidth]{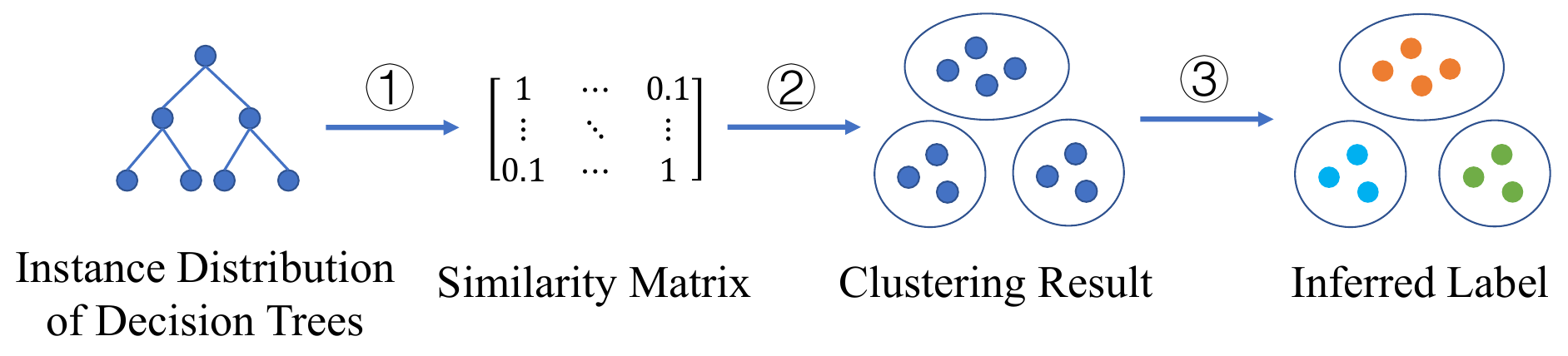}
    \vspace{-6pt}
    \caption{The workflow of instance clustering attack. 
    1) The attacker constructs a similarity matrix based on the instance distribution.
    2) The attacker clusters the training instances based on the similarity matrix.
    3) The attacker infers labels of unlabeled instances based on the known labels.
    }
    \label{fig:attack}
    \vspace{-6pt}
\end{figure}

While SecureBoost utilizes homomorphic encryption to protect gradient information passed between the active party and passive parties, it does not protect the instance distribution of each leaf node owned by the passive party, indicating that the passive party (i.e., the attacker) can exploit this information to infer the active party's labels through clustering. We name this label inference attack \textit{Instance Clustering Attack}, and its procedure is illustrated in Fig.~\ref{fig:attack} and described in Algo.~\ref{alg:attack}. 

\begin{algorithm}[!ht]\vspace{-3pt}
    \caption{Instance Clustering Attack}
	\begin{algorithmic}[1]
    \vspace{2pt}
    \Statex \textbf{Input:} Instance distribution $\{I_i\}_{i=1}^{n}$. 
    \Statex \textbf{Output:} Predicted label $\hat{y}$. 
    \vspace{4pt}
    \State {Calculate instance similarity matrix $S$ using Eq.(\ref{sim})}
    \State Categorize instances into $C$ clusters $\{c_i\}_{i=1}^C$ based on $S$.
    \For{$i=1$ to $C$}
        \State {$y_i \leftarrow$ the known label of one instance in cluster $c_i$. }
        \State {Assign label $y_i$ to $\hat{y}$ for all instances in $c_i$. }
    \EndFor
    \State \Return {$\hat{y}$}
    \end{algorithmic}\label{alg:attack}
\end{algorithm}

The attacker first constructs the similarity matrix based on the instance distribution of each leaf node using Eq.~(\ref{sim}) (line 1 of Algo.~\ref{alg:attack}).
\begin{equation}
\begin{split}
\label{sim}
sim(a, b)= & \frac{1}{n}\sum_{i=1}^{n}s(a, b, i), \\
\text{where  } s(a,b,i)= & \begin{cases}1,\text{ if } \exists j, a,b \in I_{i,j}\\ 0,\text{ otherwise} \end{cases}
\end{split}
\end{equation}
where $n$ denotes the number of decision trees, $j$ denotes the index of leaf node. 

Then, the attacker leverages this similarity matrix to categorize instances into $C$ clusters (line 2). 
We assume the attacker knows the label of one instance in each cluster. Hence, the attacker assigns instances in each cluster with the known labels (lines 3-5). 

\subsection{Defense Methods}
\label{sec:defense}

The performance of the instance clustering attack depends on the purity of leaf nodes owned by the attacker (i.e., passive party). We define the purity of a node as the ratio of instances belonging to the majority class to all instances in that node. 
A higher purity implies a more accurate similarity matrix, which can lead to more precise clustering results. 
We provide two defense methods to suppress this attack by reducing the purity: 
(1) local trees; 
(2) purity threshold.

\fakeparagraph{Local Trees}
At the first few rounds of the training, there is a strong correlation between the sign of the gradients and labels, which causes instances with the same label to be more likely to be classified into the same node, resulting in high purity of the leaf nodes. 
We propose a local training stage preceding SecureBoost, in which the active party locally trains $n_l$ decision trees, aiming to reduce the correlation between the sign of instance gradients transmitted to the passive parties and their respective labels, thereby reducing the purity of each leaf node.
\footnote{
These defense method has been implemented in FATE v1.11.2.
}

\fakeparagraph{Purity Threshold}
We apply a threshold to the node purity to prevent high-purity nodes from being disposed to the passive parties, thereby mitigating label leakage. 
When the purity of a node $j$ passes a prespecified threshold $p$, the passive party is not allowed to participate in the federated training of the subtree rooted at the node $j$, and the active party continues to train the subtree locally. 

Although the two defense methods can effectively thwart the instance clustering attack, they may introduce utility loss. Therefore, tradeoffs need to be made between utility and privacy. We consider their hyperparameters as decision variables when we optimize the Constrained Multi-Objective SecureBoost problem (Sec.~\ref{sec:cmosb}).

\section{Constrained Multi-Objective SecureBoost}\label{sec:cmosb}

In this section, we define the measurements of utility loss, training cost, and privacy leakage as well as our Constrained Multi-Objective SecureBoost algorithm.

\subsection{Measurements of Objectives}
\label{sec:objectives}

We provide measurements of the three objectives we aim to minimize: utility loss, training cost, and label leakage.

\fakeparagraph{Utility Loss}
The utility loss $\epsilon_u$ measures the amount of decrease in utility when certain defense methods are applied to thwart the label inference attack. 
\begin{equation}
    \epsilon_{u}=1-U(M, D)
\end{equation}
where $M$, $D$, and $U$ denote the SecureBoost model, the test dataset, and the performance metric, respectively. $U$ is AUC for binary classification and accuracy for multi-class classification. 

\fakeparagraph{Training Cost}
We use training time to measure the training cost of SecureBoost. Since HE operations dominate training time, we estimate the training cost $\epsilon_c$ by summing the time spent on each HE operation.
\begin{equation}
\epsilon_{c}=c_{enc}\times t_{enc}+c_{dec}\times t_{dec}+c_{add}\times t_{add}
\end{equation}
where $c_{enc}$, $c_{dec}$, and $c_{add}$ denote the number of operations for encryption, decryption, and addition, respectively; $t_{enc}$, $t_{dec}$, and $t_{add}$ denote the average time required for each HE operation, respectively. 

\fakeparagraph{Privacy Leakage}
We use the accuracy of the instance clustering attack (see Sec. \ref{sec:attack}) as the metric to measure the privacy leakage $\epsilon_p$ based on a set of instances, denoted as $I_{pl}$, randomly sampled from the training set (the number of instances belonging to each class is the same in $I_{pl}$). 
\begin{equation}
\epsilon_{p}=\frac{1}{N_{pl}}\sum_{i\in I_{pl}} \mathcal{I}[\hat{y_i}==y_i]
\end{equation}
where $\hat{y_i}$ denotes the label inferred by the instance clustering attack, and $y_i$ denotes the true label, $N_{pl}=|I_{pl}|$ and $\mathcal{I}$ denotes the indicator function. 

\subsection{Constrained Multi-Objective SecureBoost Algorithm}
\label{sec:mosb}

In this section, we propose a Constrained Multi-Objective SecureBoost algorithm (CMOSB), through which approximate Pareto optimal solutions and front can be obtained to balance the three optimization objectives and provide guidance for hyperparameter selection.

\begin{algorithm}[!ht]
	\caption{Constrained Multi-Objective SecureBoost (CMOSB)}
	\begin{algorithmic}[1]
	\Statex \textbf{Input:} Generations $T$, dataset $D_k$ owned by client $k \in [K]$, constraints $\phi_p, \phi_c$.
    \Statex \textbf{Output:} Pareto optimal solutions and Pareto front $\{X_{T}, Y_{T}\}$
    \State Initialize solutions $\{X_{0}\}$.
    \For{each generation $t$ $=1,2,\cdots,T$} 
        \State Crossover and mutate parent solutions $X_{t-1}$ to produce offspring solutions $P$; 
        \State $R$ $\leftarrow$ Merge $X_{t-1}$ and $P$;
        \State $Y$ $\leftarrow$ SBO$(R, \{D_j\}_{j=1}^K)$ 
        
        \For{each tuple ($\epsilon_p, \epsilon_c$) in $V$} 
        \State  $\epsilon_{i} = \epsilon_{i} \text{ + } \alpha_i \max\{0, \epsilon_i-\phi_i\}, i \in \{p, c\}$
        \EndFor
        \State $R^S \leftarrow$ Non-dominated sorting and crowding distance sorting $R$ based on $Y$;
        \State $X_{t}$ $\leftarrow$ Select $N$ high-ranking solutions from $R^S$;
    \EndFor
         \State \Return $\{X_{T}, Y_{T}\}$
	\end{algorithmic}\label{alg:nsga_fl}
\end{algorithm}

CMOSB is based on NSGA-II, and it is described in Algo.~\ref{alg:nsga_fl}: offspring solutions of the current generation are generated by performing crossover and mutation using the previous solution (line 3).
All newly generated individuals are evaluated using Algo.~\ref{alg:flo} (line 5). The algorithm adds constraints on the solutions (line 6) to limit the search space of the solutions.
Lines 7-8 sort the solutions and select the top $N$ individuals for the next iteration.

\begin{algorithm}[!ht]\vspace{-3pt}
    \caption{SecureBoost Optimization (SBO)}
	\begin{algorithmic}[1]
    \vspace{2pt}
    \Statex \textbf{Input:} Solutions X, feature data $D_k$ owned by client $k$.
    \Statex \textbf{Output:} Objective values $Y$ for $X$.
    \vspace{4pt}
    \For{each solution $x$ $\in$ $X$}
    
    \State{Set hyperparameters $n_l$, $n_f$, $p$ according to $x$; 
    }
    \For{each boosting round $i \in 1, \dots, n_l$}
        \State \gray{$\triangleright$ \textit{Local training stage}}
        \State {Active party train decision tree $i$ locally; }
    \EndFor
    \For{each boosting round $i \in n_l + 1, \dots, n_l + n_f$}
        \State \gray{$\triangleright$ \textit{Federated learning stage}}
        \State {Active party computes gradient $\left\langle g\right\rangle$, hessian $\left\langle h\right\rangle$; }
        \For{each node $j$ satisfies depth criteria}
            \State {Active party computes purity $p_j$ of node $j$.}
            \If{$p_j < p$}
            \State {Get instance space $I_{i, j}$ of current node; }
            \State {Split the node using $\text{SF}(I_{i, j}, \left\langle g\right\rangle, \left\langle h\right\rangle)$; }
            \Else
            \State {Active party split the node locally; }
            \EndIf
        \EndFor
        \State {Measure training cost $\epsilon_{e, i}$;}
        \State {Measure privacy leakage $\epsilon_{p, i}$;}
        \State {$\epsilon_{e} \leftarrow \epsilon_{e} + \epsilon_{e, i}$}
        \State {$\epsilon_{p} \leftarrow \max(\epsilon_{p}, \epsilon_{p, i})$}
    \EndFor
    \State {Measure utility loss $\epsilon_{u}$;}
    \State $Y \leftarrow Y + (\epsilon_p, \epsilon_c, \epsilon_u)$
    \EndFor
    \State \Return $Y$;
    \end{algorithmic}\label{alg:flo}
\end{algorithm}

Algo~\ref{alg:flo} aims to measure utility loss, training cost, and privacy leakage for training a SecureBoost model.
The solution $x$ represents a set of hyperparameters for SecureBoost, where Table~\ref{tab:variable} provides a detailed description of the hyperparameters.
Lines 3-5 correspond to the local training stage mentioned in Sec.~\ref{sec:defense}, while lines 6-19 correspond to the federated learning stage, which together constitutes the SecureBoost framework. 
In line 10, the active party calculates the purity of node $j$, and if it is below the threshold, all participants will jointly calculate the split point, otherwise, only the active party will calculate it locally.
Training cost and privacy leakage will be calculated during the iteration process, while utility will be calculated after the training is completed.

\section{Experiments}\label{sec:exp}

In this section, we empirically investigate the effectiveness of our proposed attacking method, defense methods, and the CMOSB algorithm.

\subsection{Experimental Settings}

\textbf{Dataset and setting.} 
We conduct experiments on two synthetic datasets and two real-world datasets. 
The synthetic datasets, generated using the \textit{sklearn} library\footnote{https://scikit-learn.org/stable/}, consist of Synthetic1 for binary classification and Synthetic2 for multi-class classification. 
The DefaultCredit
\footnote{https://www.kaggle.com/uciml/default-of-credit-card-clients-dataset} contains the task of predicting whether users can repay their loans on time, while the Sensorless is for sensorless drive diagnosis. 

To create datasets for the federated scenario, each dataset was vertically partitioned into two sub-datasets. 
We use 2/3 of the data as the train set and the remaining as the test set.
Table~\ref{tab:dataset} summarizes these datasets.

\begin{table}[!h]
    \centering  
    \begin{tabular}{c|c|c|c|c}  
        \hline  
        Name & S & AF & PF & C\\  
        \hline
        \hline
        Synthetic1 & 2,000 & 5 & 5 & 2\\
        \hline
        DefaultCredit & 30,000 & 12 & 13 & 2\\
        \hline
        Synthetic2 & 10,000 & 5 & 5 & 10\\
        \hline
        Sensorless & 58,509 & 12 & 36 & 11\\
        \hline
    \end{tabular}
    \caption{Datasets for evaluation. S: \# of samples; PF: \# of features in passive party; AF: \# of features in active party; C: \# of classes.}  
    \label{tab:dataset}  
    \vspace{-6pt}
\end{table}

\begin{table*}[!ht]
    \centering  
    \begin{tabular}{c|c|c|c}  
        \hline  
        Variable&Range&Chromosome Type&Description \\  
        \hline
        \hline
        $n_f$&$[1, 16]$&Binary&The number of rounds for boosting in federated learning\\
        \hline
        $n_l$&$[1, 16]$&Binary&The number of rounds for boosting locally\\
        \hline
        $d$&$[1, 8]$&Binary&Maximum depth of each decision tree \\
        \hline
        $r$&$[0.1, 1.0]$&Real-value&Subsample ratio of the training instances\\
        \hline
        $p$&$[0.1, 1.0]$&Real-value&The purity threshold to stop the federated learning\\
        \hline
        $\eta$&$[0.01, 0.3]$&Real-value&Step size shrinkage used in update to prevents overfitting\\
        \hline
    \end{tabular}
    \caption{Hyperparameter variables used in CMOSB algorithm.}  
    \label{tab:variable}  
    \vspace{-6pt}
\end{table*}

\textbf{CMOSB Setup.} 
The proposed CMOSB algorithm is based on NSGA-II. Thus, we follow the setup proposed in literature~\cite{DBLP:journals/tnn/ZhuJ20}. 
We apply a single-point crossover for binary chromosomes with a probability of 0.9 and a bit-flip mutation with a probability of 0.1. 
We apply a simulated binary crossover (SBX)~\cite{DBLP:journals/compsys/DebK95} for real-valued chromosome with a probability of 0.9 and $n_c$ = 2, and a polynomial mutation with a probability of 0.1 and $n_m$ = 20, where $n_c$ and $n_m$ denote spread factor distribution indices for crossover and mutation, respectively.
The number of generations for CMOSB is set to 40.

\subsection{Effectiveness of Defense Methods}
In this section, we investigate the efficacy of our proposed defense methods in mitigating privacy leakage.
The experiments are conducted on Synthetic1, with the default hyperparameter set as follows: $n=20$, $d=7$, $\eta=0.1$, $r=0.8$. 

\begin{figure}[!h]
    \vspace{-6pt}
    \centerline{
        \begin{minipage}[t]{0.6\linewidth}
            \centering
            \includegraphics[width=\textwidth]{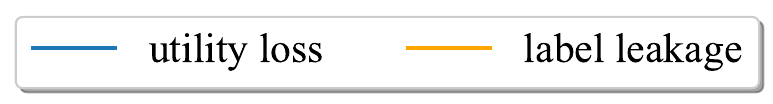}
        \end{minipage}%
    }
    \begin{minipage}[t]{0.49\linewidth}
        \centering
        \includegraphics[width=\textwidth]{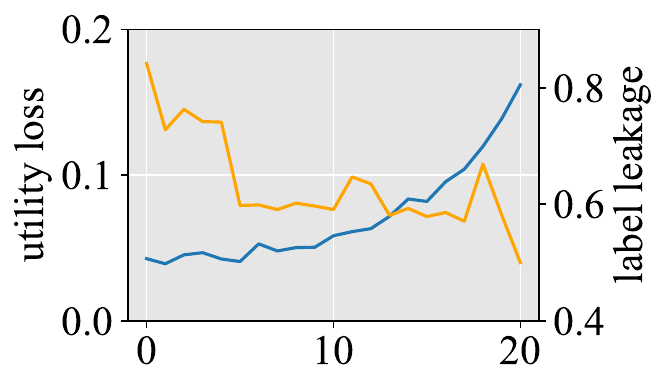}
        \centerline{(a) Varying $n_l$}
    \end{minipage}%
    \begin{minipage}[t]{0.49\linewidth}
        \centering
        \includegraphics[width=\textwidth]{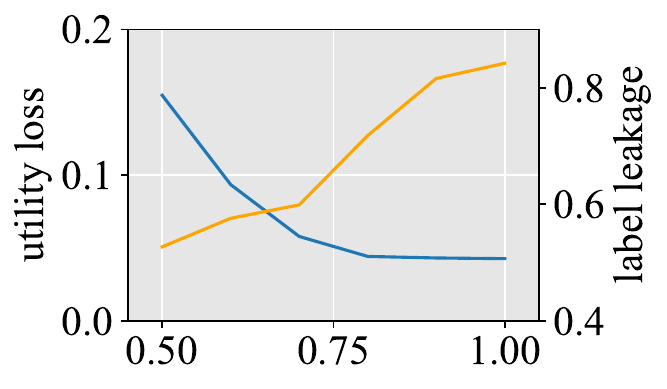}
        \centerline{(b) Varying $p$}
    \end{minipage}%
    \caption{
    Effectiveness of Defense Methods.
    Reducing $p$ or increasing $n_l$ can decrease label leakage while sacrificing utility.
    }
    \label{fig:var}
    \vspace{-6pt}
\end{figure}

Fig.~\ref{fig:var} shows that both methods can trade-off model performance for privacy protection.
As shown in Fig.~\ref{fig:var}(a), training local trees can reduce the correlation between sign of gradients and labels, thereby reducing privacy leakage of active party. 
Training 10 decision trees locally reduced label leakage risk by 25.1\% while sacrificing only 1.6\% of utility. 
The impact of purity threshold is illustrated in Fig.~\ref{fig:var}(b), where we observe a decrease in privacy leakage as more nodes are constructed by active party locally. 

\subsection{Optimal Tradeoffs between Utility, Efficiency, and Privacy achieved by CMOSB}

In this section, we use CMOSB (Algo.~\ref{alg:nsga_fl}) without using constraints to find the Pareto optimal solutions of hyperparameters for SecureBoost. Each solution is a set of hyperparameters that achieve an optimal tradeoff between utility loss, training cost, and privacy leakage. 

We summarize the variables and their descriptions involved in multi-objective optimization in Table~\ref{tab:variable}. 
Among the variables, the definitions of $n_f$, $d$, $r$, and $\eta$ are the same as those in SecureBoost. $n_l$ refers to the number of rounds for boosting locally, and $p$ refers to the threshold of purity. For binary classification problems, we further constrain the range of $p$ to $[0.7, 1.0]$ to improve the search efficiency. 

\begin{table}[!ht]
    \centering  
    \begin{tabular}{c|c|c|c|c}  
        \hline  
        Baseline&$n_f$&$d$&$\eta$&$r$\\  
        \hline
        \hline
        Fate&5&3&0.3&0.8\\
        \hline
        Emperical&10&5&0.3&0.8\\
        \hline
        VF\textsuperscript{2}Boost&20&7&0.1&0.8\\
        \hline
    \end{tabular}
    \caption{Default hyperparameters used in the comparison. } 
    \label{tab:hp}  
    \vspace{-6pt}
\end{table}

We use the default parameters of Fate~\cite{DBLP:journals/jmlr/LiuFCXY21} and VF\textsuperscript{2}Boost~\cite{DBLP:conf/sigmod/FuSYJXT021}, as well as a parameter empirically selected by us, as baselines to evaluate the effectiveness of our algorithm. 
To make a fair comparison, the sampling rate is set to $0.8$, and $complete\_secure$ is set to true. 
The specific parameter settings are shown in Table~\ref{tab:hp}.

\begin{figure}[!ht]
    \centerline{
        \begin{minipage}[t]{0.5\linewidth}
            \centering
            \includegraphics[width=\textwidth]{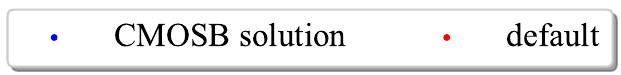}
        \end{minipage}%
    }
    \begin{minipage}[t]{0.49\linewidth}
        \centering
        \includegraphics[width=\textwidth]{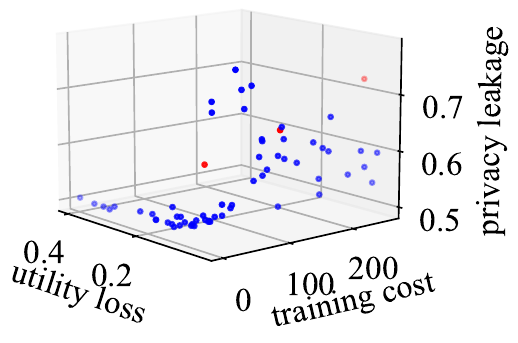}
        \centerline{(a) Synthetic1}
    \end{minipage}%
    \begin{minipage}[t]{0.49\linewidth}
        \centering
\includegraphics[width=\textwidth]{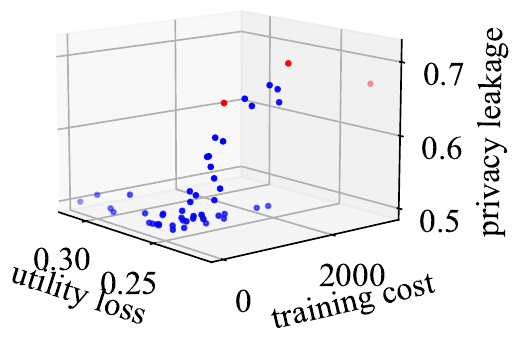}
        \centerline{(b) DefaultCredit}
    \end{minipage}
    \begin{minipage}[t]{0.49\linewidth}
        \centering
        \includegraphics[width=\textwidth]{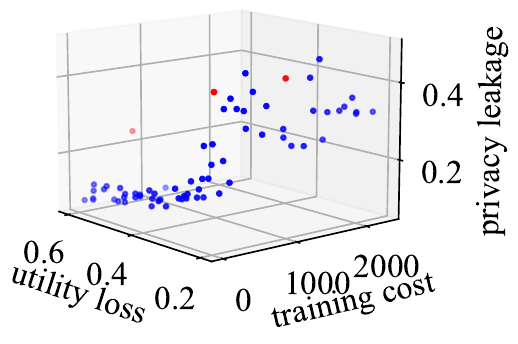}
        \centerline{(c) Synthetic2}
    \end{minipage}%
    \begin{minipage}[t]{0.49\linewidth}
        \centering
        \includegraphics[width=\textwidth]{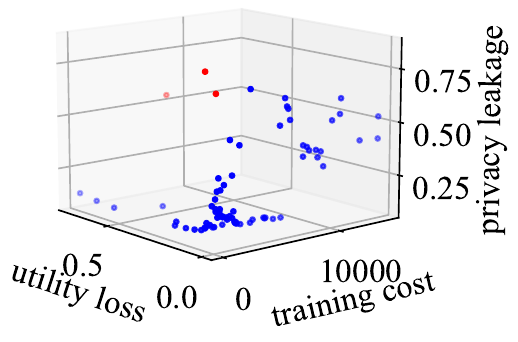}
        \centerline{(d) Sensorless}
    \end{minipage}
    \vspace{-4pt}
    \caption{
    The 3D Pareto front obtained by CMOSB.
    The blue dots represent the solutions on the Pareto front, while the red dots represent the baseline solutions. 
    Solutions closer to the origin are better. 
    }
    \label{fig:exp-3d}
    \vspace{-10pt}
\end{figure}

\begin{figure*}[!t]
    \centerline{
        \begin{minipage}[t]{0.4\linewidth}
            \centering
            \includegraphics[width=\textwidth]{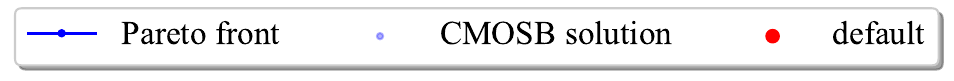}
        \end{minipage}%
    }
    \vspace{4pt}
    \begin{minipage}[t]{0.99\linewidth}
        \centering
        \includegraphics[width=\textwidth]{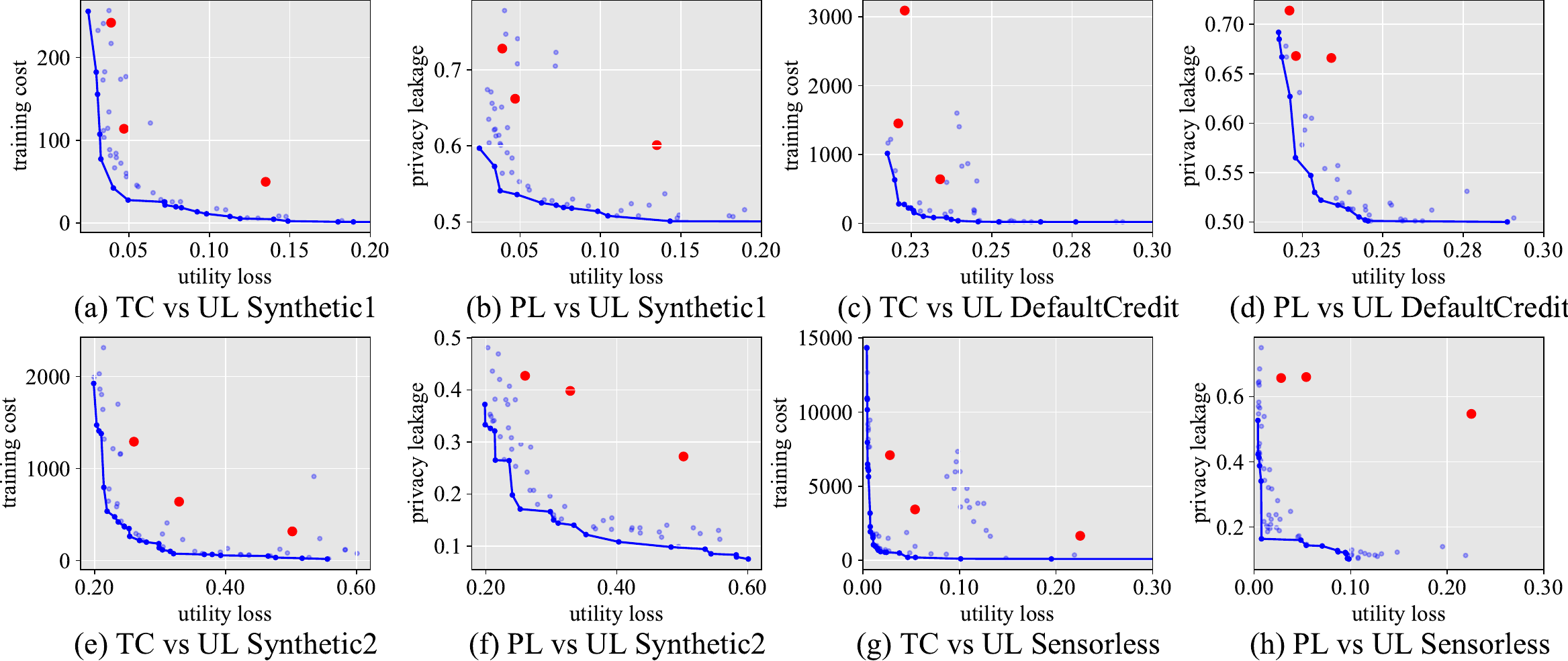}
    \end{minipage}
    \vspace{-4pt}
    \caption{
    Comparing the Pareto front of our proposed CMOSB and default hyperparameter. The first row is for two binary classification tasks, Synthetic1 and Credit2, while the second row is for multi-class classification tasks, Synthetic2 and Sensorless. In each sub-figure, the solutions closer to the bottom left corner are considered better. UL: utility loss; TC: training cost; PL: privacy leakage. 
    }
    \label{fig:exp-2d}
    \vspace{-6pt}
\end{figure*}

In the binary classification task, the 3D Pareto front is shown in Fig.~\ref{fig:exp-3d}(a-b), where the blue points represent the Pareto front that is closer to the origin than the red baseline, indicating that the solutions on the Pareto front are better than those obtained by the baseline.
It is more intuitive to observe the trade-off between utility loss, training cost, and privacy leakage in the 2D Pareto front illustrated in Fig.~\ref{fig:exp-2d}(a-d). 
For the tradeoff between training cost and utility loss (Fig.~\ref{fig:exp-2d}(a)(c)), both datasets show that as the utility loss decreases, the training cost increases continuously. This trend is more pronounced in the credit2 dataset due to its complexity.
For the tradeoff between privacy leakage and utility loss (Fig.~\ref{fig:exp-2d}(b)(d)), decreasing one would increase in the other objective. The default hyperparameters always have relatively high privacy leakage due to the privacy leakage issue mentioned in Sec.~\ref{sec:attack}.

In multi-class classification tasks, the tradeoff between the three optimization objectives also exists, as shown in the 3D Pareto front in Fig.~\ref{fig:exp-3d}(c-d).
However, the probability of privacy leakage is lower than that in binary classification tasks (Fig.~\ref{fig:exp-2d}(f)(h)), as the model becomes more complex with increasing task difficulty, making the attack more difficult. 
In Fig.~\ref{fig:exp-2d}(g)(h), when the utility loss is small, both the training cost and privacy leakage increase dramatically. This is because the task is too simple, and a good model can be trained without passive parties.

\subsection{Multi-Objective SecureBoost under Constraints}

In real-world FL scenarios, FL participants typically have requirements on objectives of federated learning. Hence, we apply the CMOSB algorithm (with constraints) to ensure that the Pareto optimal solutions found satisfy the constraints as much as possible. Adding constraints focuses the search space of the CMOSB algorithm on the feasible region, which helps find better Pareto optimal solutions more efficiently. 

\begin{figure}[!ht]
    \begin{minipage}[t]{0.49\linewidth}
        \centering
        \includegraphics[width=\textwidth]{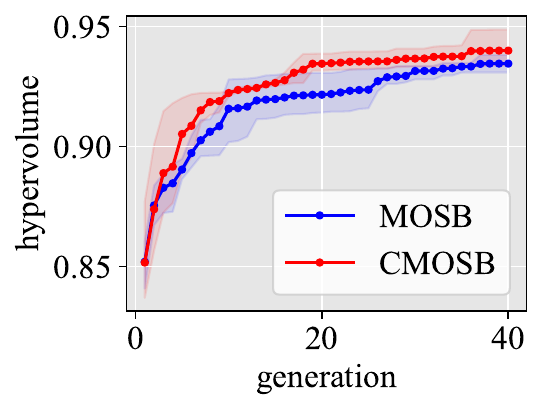}
        \centerline{(a) PL Constraint}
    \end{minipage}%
    \begin{minipage}[t]{0.51\linewidth}
        \centering
        \includegraphics[width=\textwidth]{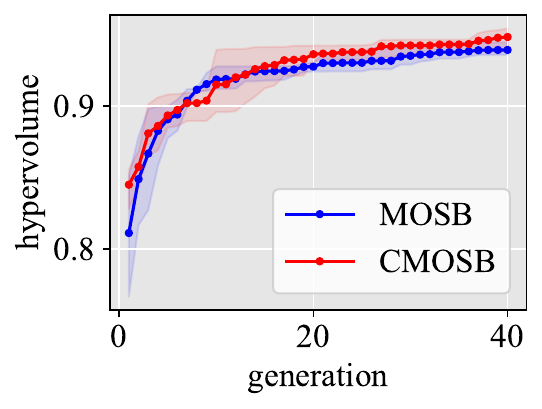}
        \centerline{(b) TC Constraint}
    \end{minipage}%
    \caption{
    Comparison of hypervolume between MOSB and CMOSB algorithms. 
    TC: training cost; PL: privacy leakage. 
    The red line represents CMOSB, and the blue line represents MOSB.
    A higher hypervolume implies that better solutions can be found. 
    }
    \label{fig:constraint-hv}
    \vspace{-6pt}
\end{figure}

We conduct this experiment on Synthetic1 and constrain the training cost and privacy leakage below 100 seconds and 0.6, respectively. We set the penalty coefficient to 20.

Fig.~\ref{fig:constraint-hv} illustrates the hypervolume comparison between CMOSB and MOSB when applying constraints to training cost and privacy leakage, respectively. 
For the privacy leakage constraint, CMOSB grows more rapidly in the first few generations and surpasses MOSB in the final generation, indicating that it can effectively find better Pareto solutions. 
For the training cost constraint, CMOSB also surpasses MOSB after 10th generation and maintains a lead to the last generation.

\begin{figure}[!h]
    \centerline{
        \begin{minipage}[t]{0.99\linewidth}
            \centering
            \includegraphics[width=\textwidth]{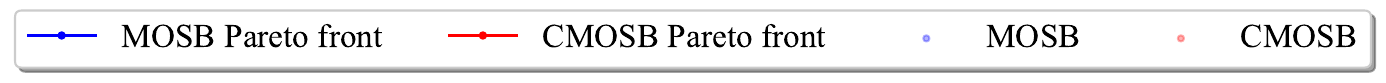}
        \end{minipage}%
    }
    \begin{minipage}[t]{0.49\linewidth}
        \centering
        \includegraphics[width=\textwidth]{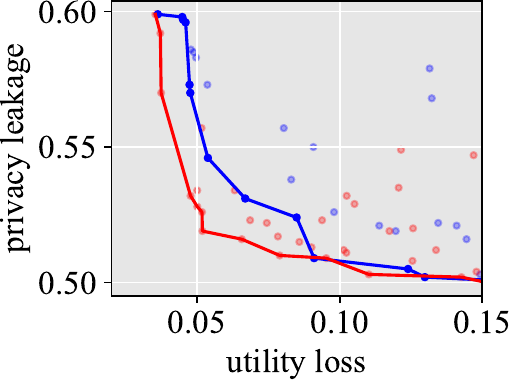}
        \centerline{(a) PL vs UL}
    \end{minipage}%
    \begin{minipage}[t]{0.51\linewidth}
        \centering
        \includegraphics[width=\textwidth]{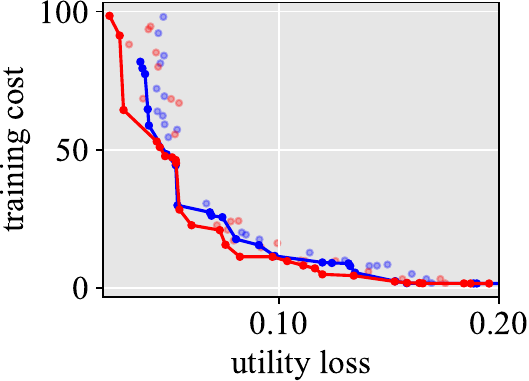}
        \centerline{(b) TC vs UL}
    \end{minipage}
    \vspace{-4pt}
    \caption{
    Comparison of Pareto front between MOSB and CMOSB algorithm. 
    UL: utility loss; TC: training cost; PL: privacy leakage.  
    We add a constraint on PL in Figure(a), and add a constraint on TC in Figure(b). 
    The red line represents CMOSB, and the blue line represents MOSB. 
    The solutions located closer to the bottom left corner of the graph are considered better.
    }
    \label{fig:constraint-2d-pareto}
    \vspace{-6pt}
\end{figure}

We further compare the Pareto fronts (at the 40th generation) obtained by the CMOSB and MOSB algorithms. 
Fig.~\ref{fig:constraint-2d-pareto}(a) demonstrates the effect of adding a privacy leakage constraint, showing a reduction of approximately 3\% in privacy leakage of the Pareto solutions at the same level of utility loss. 
The Pareto front found by CMOSB in Fig.~\ref{fig:constraint-2d-pareto}(b) is also superior to that found by MOSB, especially for solutions with lower utility loss. 

\section{Conclusion and Future Work}

In this paper, we propose the Constrained Multi-Objective SecureBoost (CMOSB) algorithm that identifies Pareto optimal solutions for SecureBoost hyperparameters by minimizing utility loss, training cost, and privacy leakage simultaneously. 
Experimental results demonstrate that the Pareto optimal solutions found by CMOSB outperform the baseline hyperparameters in terms of the tradeoff between utility loss, training cost, and privacy leakage. 
For future works, we will compare CMOSB with more sophisticated hyperparameter search algorithms, such as Bayesian Optimization, and deploy the CMOSB in production to compare its utility, efficiency, and privacy with the FATE platform. 
Furthermore, we plan to consider model complexity as one of the objectives to find model hyperparameters with higher interpretability.

\section*{Acknowledgements}

We are grateful to anonymous reviewers for their constructive comments. 
This work is partially supported by the National Science
Foundation of China (NSFC) under Grant No. U21A20516 and 62076017, the
Beihang University Basic Research Funding No. YWF-22-L-531, the Funding
No. 22-TQ23-14-ZD-01-001 and WeBank Scholars Program.

\bibliographystyle{named}
\bibliography{ijcai23}

\end{document}